# Classification of residential and non-residential buildings based on satellite data using deep learning

Jai Singla[a]* and Keivalya Pandya[b]

[a] *Space Applications Centre[1], ISRO, Ahmedabad – 380015, India*

[b] *Department of Mechanical Engineering, Birla Vishvakarma Mahavidyalaya, Vallabh Vidyanagar, India*

Corresponding Author : *jaisingla@gmail.com

# Classification of residential and non-residential buildings based on satellite data using deep learning

Accurate classification of buildings into residential and non-residential categories is crucial for urban planning, infrastructure development, population estimation and resource allocation. It is a complex job to carry out automatic classification of residential and non-residential buildings manually using satellite data. In this paper, we are proposing a novel deep learning approach that combines high-resolution satellite data (50 cm resolution Image + 1m grid interval DEM) and vector data to achieve high-performance building classification. Our architecture leverages LeakyReLU and ReLU activations to capture nonlinearities in the data and employs feature-engineering techniques to eliminate highly correlated features, resulting in improved computational efficiency. Experimental results on a large-scale dataset demonstrate the effectiveness of our model, achieving an impressive overall F1-score of 0.9936. The proposed approach offers a scalable and accurate solution for building classification, enabling informed decision-making in urban planning and resource allocation. This research contributes to the field of urban analysis by providing a valuable tool for understanding the built environment and optimizing resource utilization.

1. **Introduction**

Until now majority of Indian population stayed in villages and was depending majorly on agriculture. However, slowly Indian population is making a move towards tier-I/II cities. Therefore, urban cities are expanding at a rapid pace in India. With the availability of high-resolution satellite data and rising urbanization in the country, there is a scope and opportunities to study and comprehend urban landscape in a focused manner. Urban analysis also includes the classifications of buildings into several categories (residential / commercial etc) according to their intended use. Manual classification using the data is very tedious task. Hence, automatic approach of deep learning is sought to classify buildings as residential or non-residential. Primary objective of this research paper is to devise a novel method for classification of the buildings using satellite data into residential and non-residential categories using deep learning techniques. Prominent challenges/ gaps in classifying a building in to the categories of residential / non-residential are:

i. **Complexity of Urban Environments:** Urban areas are highly heterogeneous, with buildings varying in size, shape, height, and purpose. This complexity makes it difficult to classify buildings accurately using conventional methods.
ii. **High Correlation of Features:** Many features extracted from satellite images, such as building footprints, texture, and spectral information, are highly correlated.
iii. **Limited Use of Elevation Data**: While geometric and contextual features are commonly used, the inclusion of terrain and elevation data, which can significantly influence building characteristics, is often overlooked.

The key contribution of this research paper is the proposal of an innovative deep learning architecture tailored for building classification and unique approach to classify building as residential or non-residential using satellite data. The proposed model takes advantage of both raster data, specifically the Digital Elevation Model (DEM), high resolutions satellite images, and vector data, represented by building footprint shape files. By combining these data sources, we are able to extract comprehensive features that capture both geometric and contextual information about buildings. The deep learning model consists of multiple layers, including input, hidden, and output layers, which are carefully designed to learn and discriminate the unique characteristics of residential and non-residential buildings.

## 2. Literature Survey

Several researchers have explored machine learning techniques for the classification of buildings into residential and non-residential categories in the past. One such study by Atwal et al. (2022) employed a C4.5 binary decision tree classifier with 148 nodes, including 72 leaf nodes, to classify buildings based on open street maps of Fairfax, Mecklenburg, and Boulder City. Their model achieved an accuracy of 98% for Fairfax and 96% and 93% when transferred to Mecklenburg and Boulder, respectively. While the model exhibited high accuracy in predicting residential buildings with an F1-score of 98%, its performance in predicting non-residential buildings was suboptimal, with F1-scores ranging from 74.81% to 83.99% across the study regions. Bandam, Abhilash, et al. (2022) proposed a method for predicting building types in Germany using data-driven modeling techniques by utilizing OSM data . The method involves extracting features from building footprints and other geospatial data using spatial analysis and machine learning techniques and then using these features to train a Random Forest classifier to predict building types. The paper shows that the proposed method outperforms a baseline method that uses only building area as a feature, achieving an accuracy of 88.9%, a precision of 88.8%, a recall of 88.3%, and an F1-score of 88.5%. Wei Chen, et al. (2020) proposed a framework to derive information on building types using geospatial data from Gaode and Baidu Maps, including point-of-interest (POI) data, building footprints, etc. The proposed framework uses natural language processing (NLP)-based approaches to reclassify POI categories and identify building types using two indicators of type ratio and area ratio. The framework was tested on over 440,000 building footprints in Beijing, achieving overall accuracies of 89.0% and 78.2% for NLP-based approaches and building type identification methods, respectively. Lloyd, et al. (2020) explores the use of satellite image-derived building footprint data to classify the residential status of urban buildings in low and middle-income countries. The authors applied an ensemble machine learning building classification model for the first time to the Democratic Republic of the Congo and Nigeria. The authors described a GIS workflow that semi-automated data preparation for input to the model, making it useful for those applying the model to additional countries and using input data from diverse sources. The results showed that the ensemble model correctly classifies between 85% and 93% of structures as residential and non-residential across both countries. Bing Xia, et al., (2020) proposed a framework for analyzing, comparing, and predicting the style of residential buildings based on machine learning techniques and use a dataset of 1000 residential buildings from different regions and periods in China and extract 18 features related to their shape, color, texture, and ornamentation. They applied various machine learning algorithms, such as k-means clustering, support vector machine (SVM), random forest (RF), and artificial neural network (ANN), to classify and predict the style of the buildings. Their ANN achieved an accuracy of 77.2%.

**Innovation in Deep Learning for Building Classification of residential / non-residential** , Our work highlights the following:

(1) Introducing DEM feature for modeling parameters: Traditionally, building classification models have primarily relied on geometric and contextual features derived from satellite imagery, such as building footprints, texture analysis, and spectral information. While these features have proven useful, they may not fully capture the influence of terrain and elevation on building characteristics. In our work, we propose the inclusion of DEM as an additional feature to high-resolution satellite imagery in the modeling process. DEM provides detailed information about the elevation and topography of the land surface, offering valuable insights into the physical characteristics of buildings and their surroundings.

(2) All the exiting studies mainly relied on OSM data for model training and validations purpose. However, in our study we used an earlier work Singla et al (2022) to automatically extract accurate building shape files from high-resolution satellite data. Further, refined residential/non-residential attributes are appended to each building shape by validating these shapes from multiple sources for example Google maps, Bings etc.

(3) **Proposed an ANN Architecture**: We recognized the importance of incorporating both spatial and attribute information in accurately differentiating between residential and non-residential buildings. To achieve this, we leverage the DEM raster file, which provides elevation data and captures the terrain characteristics of the area, and the building footprint shape file. By integrating this data into our ANN architecture, we enable the network to learn and extract relevant features related to building footprints. Furthermore, we utilize the building footprint shapefile, which contains geometric and attribute information of individual buildings, including their size, shape, and location. This data source enriches our feature extraction process, allowing the ANN to capture important characteristics specific to residential and non-residential buildings. Our proposed ANN architecture is designed to process and analyze these combined data sources effectively. By incorporating the DEM raster file and the building footprint shapefile, we aim to improve the accuracy and robustness of our classification model. The ANN learns and leverages the spatial relationships and attributes present in the data to make informed decisions about the building type.

(4) Highest accuracy and F1 score amongst all other previously known work: In our study, we have extensively evaluated the performance of our proposed ANN architecture on a comprehensive dataset. Through rigorous experimentation and analysis, our model has demonstrated remarkable accuracy and achieved the highest F1 score among all existing methods. The superior accuracy and F1 score achieved by our proposed ANN architecture can have significant implications in various domains. It offers more reliable information for urban planning, real estate analysis, and decision-making processes. Stakeholders and policymakers can rely on the accuracy of our model's predictions to make informed choices and optimize resource allocation.

### 3. Dataset Details

There is a requirement of high-resolution satellite imagery and high resolution Digital Elevation Model (DEM) to proceed for accurate classification of buildings in to residential vs non- residential. We made use of very high-resolution 50cm image data of World View (WV) -3 satellite as an input for this task. Apart from imagery, we also used 1m sampling DEM, derived from WV-3 satellite and supplied by AW3D team, to prepare an advanced training dataset. WV-3 is high-resolution commercial satellite launched by Digital Globe in the United States. WV-3 provides panchromatic imagery with resolution of 0.34m and multispectral resolution of 1.24m. Brief description of the data is mentioned in Table-1.

Table 1. Brief description of the data

| Satellite - Scene | Spatial resolution | Dimensions | Site | Date of acquisition/ processing |
|---|---|---|---|---|
| WV-3 | 50cm | 14173x16892 | Gandhinagar | April 2019 |
| WV-3 | 1m (RMSE ~3m) | 7088x 8446 | Gandhinagar | April 2019 |

Vector shape files: Using high-resolution images and our earlier work on automatic extraction of shapefiles from satellite data (Singla et al, 2023), we have accurately determined the shapefiles over study area. Further, attribute of residential or non-residential is appended in the shapefile table manually by carefully looking at the building parameters from satellite data as well as using ground information.

### 3.1 Data Pre-processing and Exploratory Data Analysis (EDA)

We have used Digital Elevation Model (DEM) Raster file (.tiff) and Building footprint shapefile (.shp) to extract building features. We initially map an adequate amount of heterogeneous building types in the data (such as apartments, churches, school, college, offices, etc.) to these two classes (residential / non-residential) since our objective is to predict residential and non-residential building types.

One of the key components of our data preprocessing was the creation of a derived attribute table (Table-2). Attribute table was constructed by combining the information extracted from the high-resolution images and DEM Raster files to collect final data in the shapefile format. The derived attributes included various features such as building height from DEM, and area, shape characteristics, and contextual information from high resolution image. By combining these data sources and constructing the derived attribute table, we were able to create a comprehensive set of features that captured both visual and spatial characteristics of the buildings. This enriched dataset served as the foundation for our subsequent analysis and modelling tasks, ultimately facilitating the accurate classification of residential and non-residential building types.

As described above, the attribute table (Table-2) generation process involves several critical steps aimed at integrating and synthesizing data from high-resolution images, DEM raster files, and vector shapefiles to extract and compile relevant building features. UID is the unique identifier for each building. BuildType Categorical variable indicating the building type (residential/non-residential). RoofColor is a categorical variable for roof color. Mean, Max and Std are the continuous variables representing statistical metrics related to building characteristics. Floor is a continuous variable for the number of floors. Area_sqft and Area_sqm are continuous variables for the building area in square feet and square meters. Nodes are continuous variable indicating the number of nodes (or edges) in the building's footprint. Res is the categorical variable representing residential status. Ht, is the continuous variable for building height. Geometry is the polygon representing the building's footprint.

Table 2. Derived attribute table from high-resolution images and DEM

|            | Feature type | Mean      | Max         | Std       | DType     |
|------------|--------------|-----------|-------------|-----------|-----------|
| **UID**       | Unique ID    | -         | -           | -         | object    |
| **BuildType** | Categorical  | -         | -           | -         | object    |
| **RoofColor** | Categorical  | -         | -           | -         | object    |
| **_mean**     | Continuous   | 25.2176   | 78.7562     | 3.3335    | float64   |
| **Floor**     | Continuous   | 8.4058    | 26.2520     | 1.1111    | float64   |
| **Area_sqft** | Continuous   | 1552.0230 | 191268.2879 | 4037.1331 | float64   |
| **Area_sqm**  | Continuous   | 144.1876  | 17769.4054  | 375.0619  | float64   |
| **nodes**     | Continuous   | 5.1131    | 106.0000    | 3.5721    | float64   |
| **res**       | Categorical  | -         | -           | -         | float64   |
| **ht**        | Continuous   | 7.7176    | 61.2562     | 3.3335    | float64   |
| **geometry**  | Polygon      | -         | -           | -         | object(3) |

For the data pre-processing we have used Geopandas (Geopanda library), which converts the input file into a GeoDataFrame. Training data is a crucial component in the development and performance of deep learning (DL) models. For the accurate classification of buildings into residential and non-residential categories using satellite data, our study involved meticulous preparation of the training data. Our training dataset contains details of 15,999 buildings out of which 15,582 buildings are residential accounting for 97.39% of the dataset, and 417 are non-residential buildings accounting for 2.61% of the dataset. We have divided this entire dataset as 80% for training, 10% for testing and another 10% for validations.

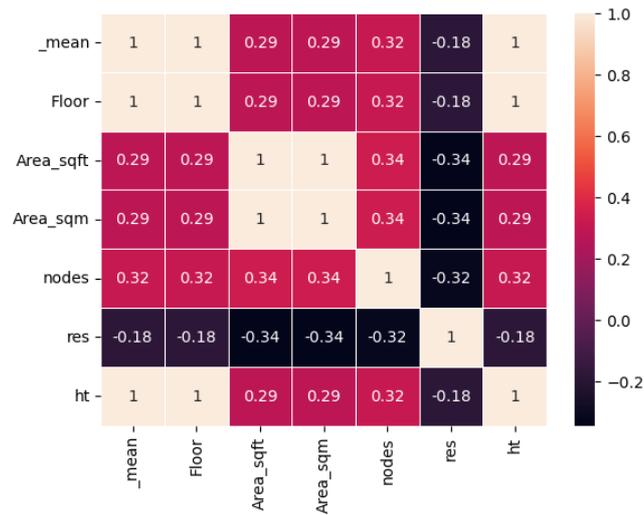

Figure 1. Correlation matrix of features of dataset

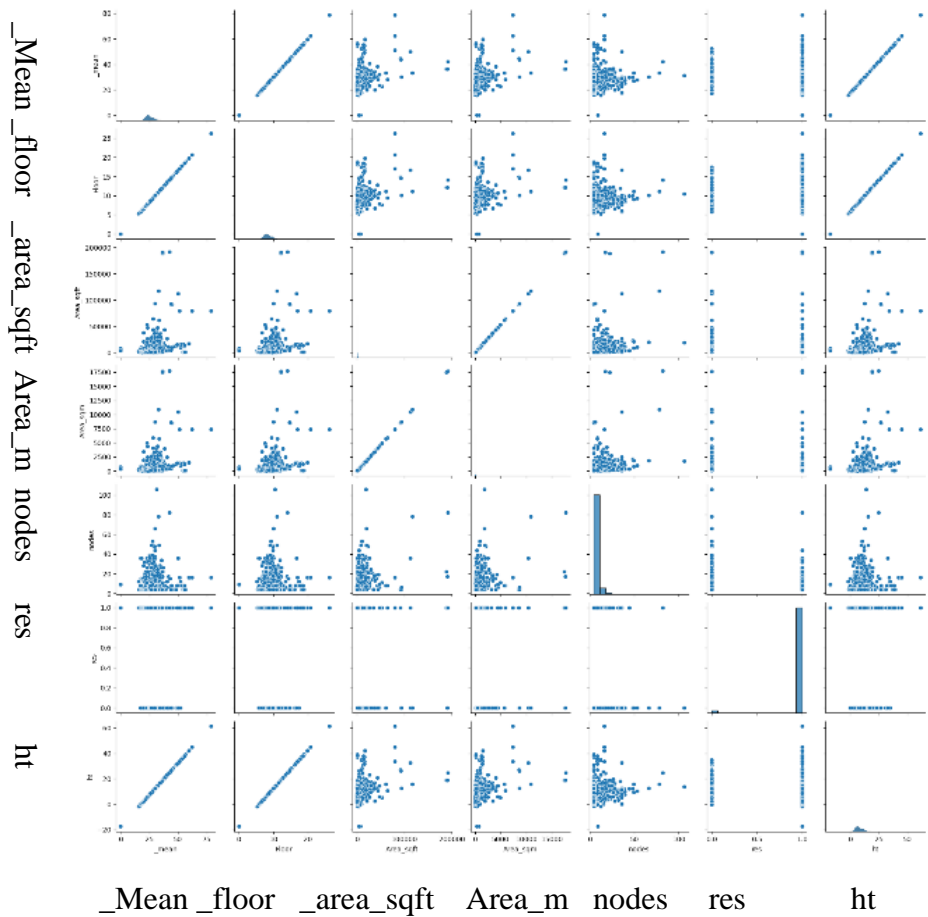

Figure 2. Pairwise scatter plot of dataset

In the process of feature engineering, we aim to improve the quality and efficiency of our dataset by making necessary modifications. One important step in this process is to address the issue of highly correlated features. In our analysis, we identified several features that exhibited high correlation, such as "_mean," "floor," "ht," "Area_sqft," and "Area_m." To mitigate redundancy and improve computational efficiency, we decided to drop the highly correlated features while retaining the most relevant ones. Specifically, we kept the features "height" and "Area-in-sqft" as they provided valuable information for our analysis. To illustrate the impact of this preprocessing step, we generated correlation matrices (Fig. 1 and Fig. 2) before and after the data preprocessing. By comparing these matrices, we can observe the changes in the correlation between features.

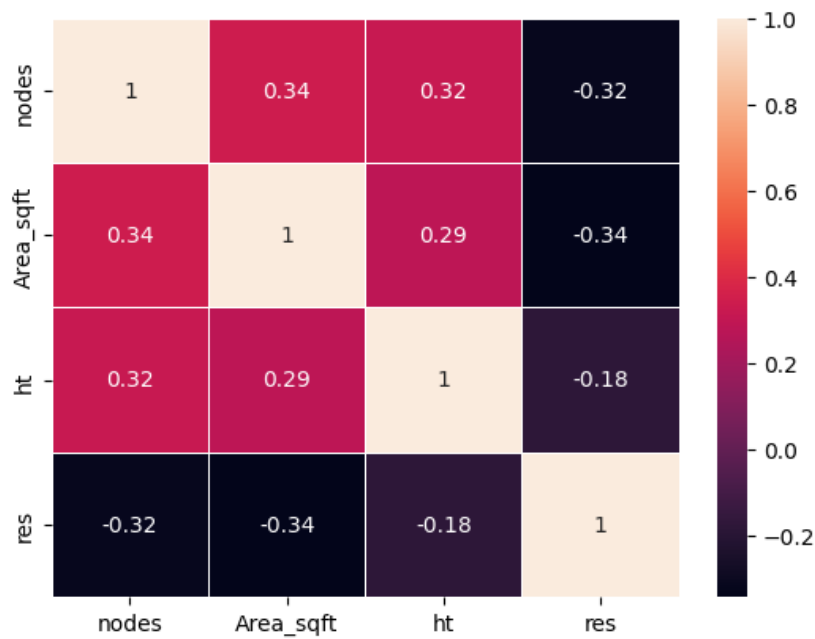

Figure 3. Correlation matrix of features after feature extraction

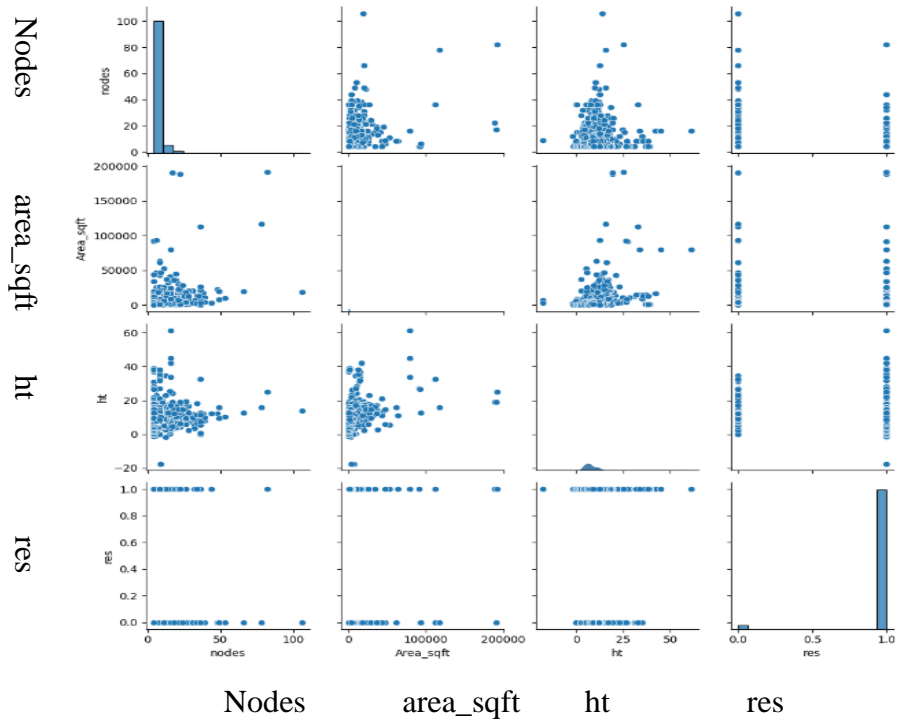

Figure 4. Pairwise scatter plot of features after feature extraction

The correlation matrix before preprocessing shows the initial relationships between all the features, including the highly correlated ones. However, after removing the redundant features, the correlation matrix after pre-processing (Fig. 3 and Fig. 4) reflects a more streamlined and concise representation of the inter-feature relationships.

This process of dropping highly correlated features not only reduces redundancy but also improves the computational efficiency of our analysis without compromising the integrity of the dataset. It allows us to focus on the most informative and independent features, thereby enhancing the accuracy and interpretability of our subsequent analyses and models.

## 4. Methodology

In this work, we are proposing a deep neural network architecture to classify building types into residential and non-residential classes. Overall flow methodology includes data preparation (section 3), defining appropriate neural network architecture, its optimization function, hyper-parameters and evaluation metrics (section 4) and generation of the results (section 5). The detailed neural network architecture, optimization function, hyper-parameters and evaluation metrics are explained as follows.

*4.1 Neural Network Architecture*

As shown in Figure-5, the first layer, named Input_Layer is a dense layer with 1024 neurons uses the LeakyReLU (Mass et. al 2013) activation function with a small negative slope (α = 0.01) to tackle "dying ReLU" problem, where a large portion of the ReLU neurons become inactive or "dead" during training, resulting into outputting zero. This occurs when the weighted sum of all the inputs to ReLU (Fukushima, 1975) is negative, causing it to output zero and effectively "kill" the neuron.

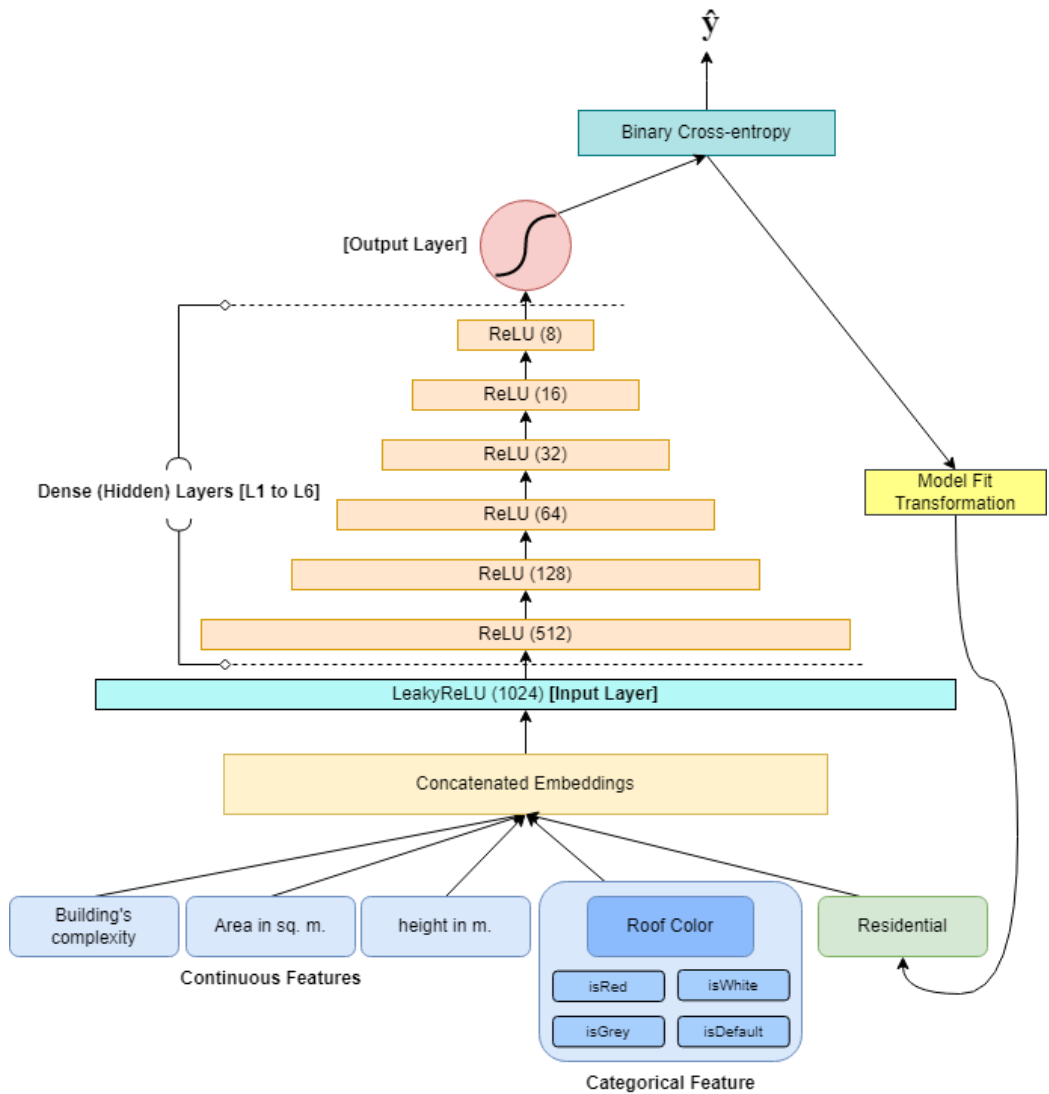

Figure 5. Proposed Neural Network Architecture

Following the input layer, six additional dense layers (Dense_L1 to Dense_L6) gradually decrease the number of neurons. ReLU is used in these hidden layers to boost computational efficiency. It sets negative values to zero, which helps in capturing non-linear relationships and introducing sparsity in the network. This sparsity assists in feature selection by reducing the impact of less informative features. The number of neurons in each layer is as follows: Dense_L1 (512 neurons), Dense_L2 (128 neurons), Dense_L3 (64 neurons), Dense_L4 (32 neurons), Dense_L5 (16 neurons), and Dense_L6 (8 neurons)

The final layer, named Output_Layer is a dense layer with a single neuron with Sigmoid as the output activation with binary cross-entropy loss function for this problem is binary in nature. It also provides a smooth and interpretable output, allowing for easy interpretation of the predicted probabilities.

The mathematical equation corresponding to the model is:

$$\text{LeakyReLU}(x) = \max(0.01x, x)$$

$$\text{ReLU}(x) = \max(0, x)$$

$$\sigma(x) = \frac{1}{1 + e^x}$$

$$Z_1 = X \cdot W_1 + b_1 \qquad A_1 = \text{LeakyReLU}(Z_1)$$

$$Z_2 = A_1 \cdot W_2 + b_2 \qquad A_2 = \text{ReLU}(Z_2)$$

$$Z_3 = A_2 \cdot W_3 + b_3 \qquad A_3 = \text{ReLU}(Z_3)$$

$$Z_4 = A_3 \cdot W_4 + b_4 \qquad A_4 = \text{ReLU}(Z_4)$$

$$Z_5 = A_4 \cdot W_5 + b_5 \qquad A_5 = \text{ReLU}(Z_5)$$

$$Z_6 = A_5 \cdot W_6 + b_6 \qquad A_6 = \text{ReLU}(Z_6)$$

$$Z_7 = A_6 \cdot W_7 + b_7$$

$$\hat{y} = \sigma(Z_7)$$

where X is the input matrix of shape (15999, 7), W1, W2, ..., W7 are the weight matrices of the layers, b1, b2, ..., b7 are the bias vectors of the layers, and ŷ is the predicted output vector of shape (15999, 1).

## *4.2 Optimization Algorithm*

The model is compiled with the Adam optimizer, using the AMSGrad variant. Adam (Adaptive Moment Estimation) is an optimization algorithm (Kingma et al, 2014), which combines the advantages of AdaGrad and RMSProp and calculates adaptive learning rates for each parameter. However, Adam algorithm can exhibit overshooting behaviour, where the learning rate adapts too aggressively and prevents the model from converging to the optimal solution. AMSGrad variant further improves convergence as it prevents overshooting by introducing a modification in the update rule (Duchi et al 2011). AMSGrad is also known for its robustness to different optimizer configurations, providing improved convergence and better generalization performance. It performs well across a wide range of deep learning tasks and architectures, making it a reliable choice for optimizing neural networks. Several empirical studies (Reddi, 2019; Wilson, 2017; Meity, 2017) have also shown that the AMSGrad can give improved convergence and better generalization performance compared to other optimization algorithms. Additionally, the AMSGrad variant achieves this by maintaining a long-term memory of past gradients, which helps in stabilizing the learning process. This is done by modifying the exponential moving average of past squared gradients, ensuring that the learning rate does not increase abruptly. By doing so, AMSGrad effectively mitigates the issue of overshooting, leading to a more stable and consistent convergence trajectory. AMSGrad's robustness and adaptability make it particularly effective in scenarios involving complex, high-dimensional data and intricate neural network architectures. For instance, in natural language processing (NLP) tasks, where models like transformers require careful tuning of optimization parameters, AMSGrad helps in achieving faster convergence without compromising on the model's generalization capabilities.

*4.3 Evaluation Metrics and Loss Function*

In order to objectively assess our model, we first gauge the neural network's precision. However, given that the dataset shows a considerable class imbalance, using accuracy alone as a criterion for evaluation might be quite deceptive. For example, an arbitrary classification method that labels all structures as residential would already have an accuracy of

$$accuracy = \frac{15582}{15582 + 417} = 0.9739$$

The ability of our algorithm to forecast non-residential buildings is an important aspect that cannot be inferred from the accuracy measures. In order to forecast the two classes, we thus compute the F1-score, which is the harmonic mean of the model's accuracy and recall.

$$F1 = 2 * (precision * recall) / (precision + recall)$$

The binary cross-entropy loss, which has a logarithmic scale, penalizes confident and inaccurate predictions more severely, which encourages the model to be more cautious and accurate in its forecasts. By giving each class equal weight during training, the logarithmic scale helps manage the dataset's imbalance—residential structures are far more common than non-residential buildings. As a result, the model is optimized to reduce the binary cross-entropy loss by producing probabilities that appropriately represent the likelihood of each class. Hence, to assess the model's effectiveness, we use the F1 Score.

During training, early stopping is employed with the "val_get_f1", a validation F1 score metric as the monitoring criterion. This helps prevent overfitting by stopping the training process if the validation F1 score does not improve for a specified number of epochs (here, patience = 50). The EarlyStopping callback is used to monitor the validation F1 score and restore the weights of the best-performing model.

*4.4 Hyperparameters*

Table-3 enlists essential hyperparameters and corresponding values, which are used in the proposed architecture.

Table-3 (Summary of Hyper-parameters)

| Hyper parameter | Value |
| --- | --- |
| Learning Rate | 0.001 |
| Activation Function | LeakyReLU ($\alpha = 0.1$), ReLU, Sigmoid |
| Objective Function | Adam with AMSGrad |
| Loss function | Binary Cross-entropy |
| Batch size | 8 |
| Epochs | 500 (and early-stopping) |
| Early-stopping Patience | 50 |

5. Results

The results of the study demonstrate the effectiveness of the developed ANN model in accurately distinguishing between residential and non-residential buildings. As seen in Table-4, with an overall F1-score of 0.9936, the model exhibits high precision and recall for both classes. Specifically, it achieves a precision of 0.99 and recall of 0.99 for the residential class, and a precision of 0.79 and recall of 0.73 for the non-residential class. These metrics reflect the model's ability to accurately predict and classify buildings based on their intended usage. Figure-6 depicts the glimpses of corrected predicted and wrongly predicted results over the study area. Figure-7 depicts the model recall and loss values during training and validation phase whereas Figure-8 contains the details about confusion matrix.

Table-4: Results and Accuracies

|                  | Precision | Recall | F1-Score | Support |
|------------------|-----------|--------|----------|---------|
| **Non-residential** | 0.79      | 0.73   | 0.76     | 84      |
| **Residential**     | 0.99      | 0.99   | 0.99     | 3116    |
| **Accuracy**        |           |        | 0.99     | 3200    |
| **Macro Average**   | 0.89      | 0.86   | 0.88     | 3200    |
| **Weighted Average**| 0.99      | 0.99   | 0.99     | 3200    |

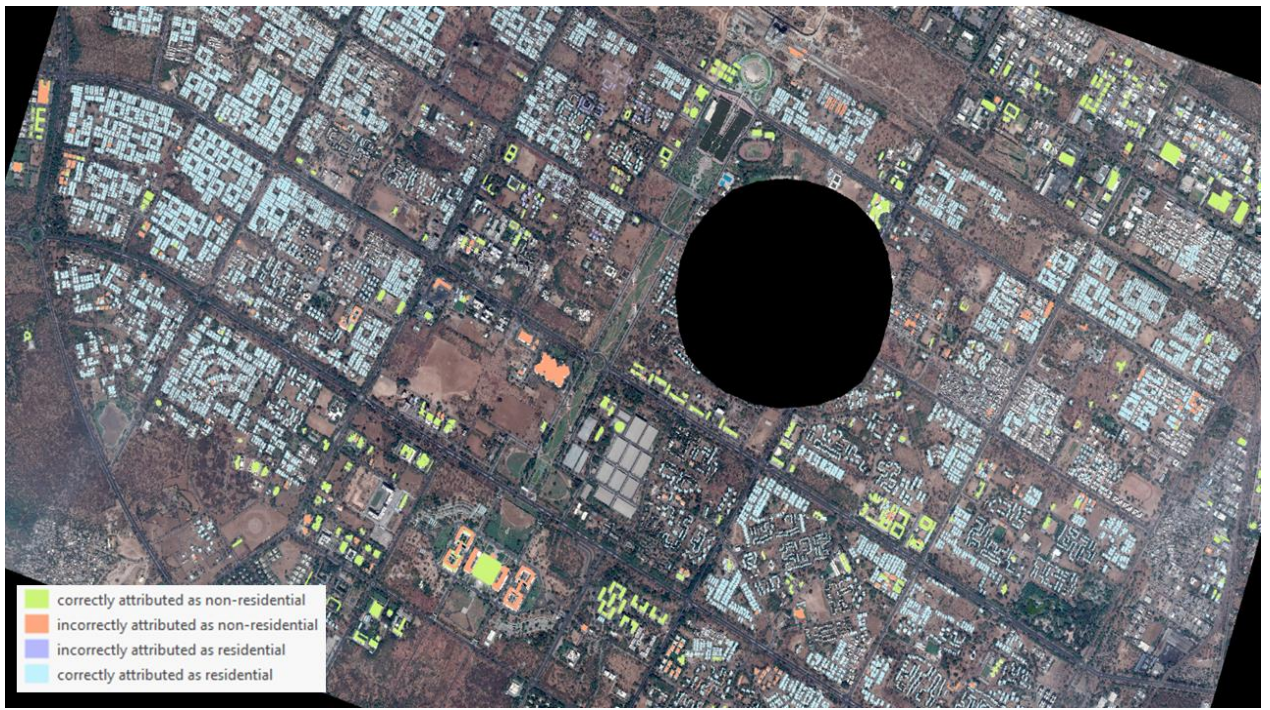

Figure 6. Residential and non-residential building type based on our proposed model for Gandhinagar, India

The overall loss value achieved by the model is 0.0391. This metric reflects the performance of the model in minimizing the discrepancy between the predicted and actual labels during the training process. The final overall F1-Score attained by the model is 0.9936.

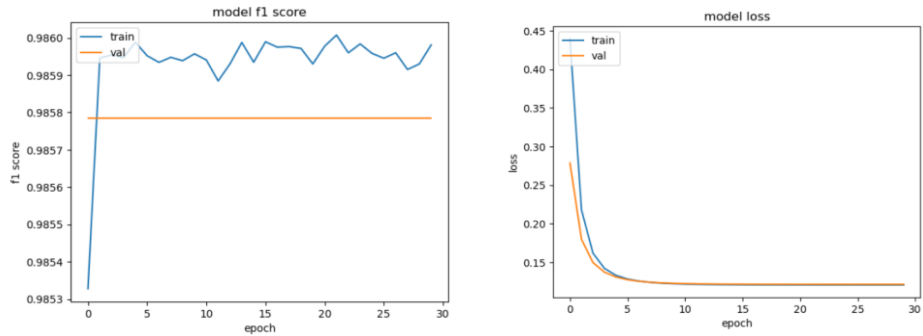

Figure 7. Model F1 Score and Loss during training and validation

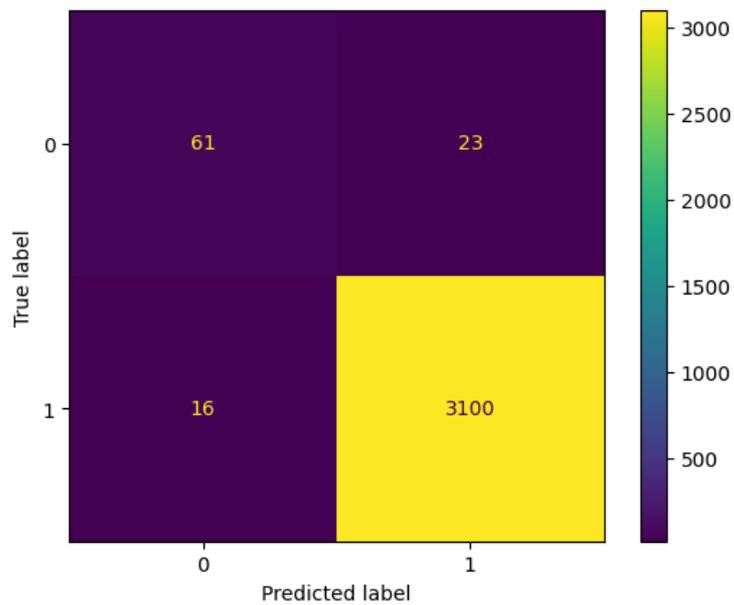

Figure 8. Confusion Matrix

## 6. Discussions

The research paper presents a novel approach for classifying buildings into residential and non-residential categories using a custom deep learning artificial neural network (ANN). This study addresses the need for accurate and automated identification of building types, which has significant implications for urban planning and decision-making processes. By leveraging building footprints from Digital Elevation Model (DEM) raster files and shapefiles, the proposed ANN architecture extracts essential building features for classification.

The significance of this research lies in its potential applications in urban planning and decision-making processes. Accurate identification of building types can aid in various urban development tasks, such as infrastructure planning, zoning regulations, and resource allocation. The proposed approach offers several advantages, including the utilization of readily available geospatial data sources and the capture of complex relationships between building features and their corresponding usage types through a custom deep learning ANN.

However, the study acknowledges certain limitations, including the need for further validation across diverse geographic regions and datasets. Additionally, considerations for potential biases and limitations in the input data should be addressed to ensure equitable and unbiased classification results. There is erroneous observations due intermixing of commercial activities in the residential areas. However, these type of issues can be fully handled using thorough ground surveys.

Despite these limitations, the developed ANN model trained on DEM raster files and shapefiles provides a promising tool for urban planners and policymakers to efficiently analyse and utilize building data. Overall, this research paper contributes to the field of geospatial analysis and urban planning by presenting a novel approach for classifying buildings based on their usage. The findings of this study lay the groundwork for further research and applications in urban planning, resource management, and decision-making processes.

## 7. Conclusion and Future Work

In conclusion, our research presents a custom deep learning artificial neural network (ANN) architecture that effectively classifies buildings into residential and non-residential categories based on their footprints. With an overall F1-score of 0.9936 and high precision and recall for both classes, our model demonstrates its potential for accurate building type identification. This study holds significant implications for urban planning, as it enables precise analysis of building data for tasks such as infrastructure planning, zoning regulations, and resource allocation. By leveraging geospatial data sources and capturing complex relationships between building features and usage types, our approach contributes to the field of geospatial analysis and provides a foundation for future research and practical applications in urban development.

In future, there are several avenues to explore for further enhancement and application of our research. Firstly, incorporating additional geospatial data such as street network data, and demographic information can provide a more comprehensive understanding of building types and their surrounding context. Furthermore, exploring the potential of transfer learning and ensemble techniques can improve the generalization and robustness of the model across different regions and datasets. Additionally, conducting extensive validation and testing on diverse/ congested urban environments and datasets would further validate the effectiveness and reliability of our approach. Finally, integrating real-time data sources and continuous model retraining can enable the dynamic monitoring and classification of building types, facilitating up-to-date urban planning and decision-making processes. Overall, these future directions would contribute to the advancement of building classification techniques and their practical applications in urban management and development.